\documentclass[conference]{IEEEtran}
\IEEEoverridecommandlockouts
\usepackage{cite}
\usepackage{amsmath,amssymb,amsfonts}
\usepackage{algorithmic}
\usepackage{graphicx}
\usepackage{textcomp}
\usepackage{xcolor}
\def\BibTeX{{\rm B\kern-.05em{\sc i\kern-.025em b}\kern-.08em
    T\kern-.1667em\lower.7ex\hbox{E}\kern-.125emX}}

\usepackage{times}
\usepackage{epsfig}
\usepackage{graphicx}
\usepackage{amsmath}
\usepackage{amssymb}

\usepackage{subfigure}
\usepackage{multirow}
\usepackage{diagbox}
\usepackage{booktabs}
\usepackage{arydshln}
\usepackage[normalem]{ulem}
\useunder{\uline}{\ul}{}
\usepackage[misc]{ifsym}

\usepackage{amsthm, amsmath , amssymb}
\usepackage{mathrsfs}

\let\OLDthebibliography\thebibliography
\renewcommand\thebibliography[1]{
  \OLDthebibliography{#1}
  \setlength{\parskip}{0pt}
  \setlength{\itemsep}{0pt plus 0.3ex}
}

\begin{document}

\title{Towards Discriminative Representations with Contrastive Instances for Real-Time UAV Tracking\\
}

\author{\IEEEauthorblockN{1\textsuperscript{st} Dan Zeng}
\IEEEauthorblockA{\textit{Southern University of Science and Technology} \\
Shenzhen, China \\
zengd@sustech.edu.cn}\\
\IEEEauthorblockN{3\textsuperscript{rd} Xucheng Wang}
\IEEEauthorblockA{\textit{Guilin University of Technology} \\
Guilin, China \\
xcwang@glut.edu.cn}
\and
\IEEEauthorblockN{2\textsuperscript{nd} Mingliang Zou}
\IEEEauthorblockA{\textit{Guilin University of Technology} \\
Guilin,China \\
393038726@qq.com}\\
\IEEEauthorblockN{4\textsuperscript{th} Shuiwang Li {\textrm\Letter}}
\IEEEauthorblockA{\textit{Guilin University of Technology} \\
Guilin, China \\
lishuiwang0721@163.com}
}

\maketitle

\begin{abstract}
Maintaining high efficiency and high precision are two fundamental challenges in UAV tracking due to the constraints of computing resources, battery capacity, and UAV maximum load. Discriminative correlation filters (DCF)-based trackers can yield high efficiency on a single CPU but with inferior precision. Lightweight Deep learning (DL)-based trackers can achieve a good balance between efficiency and precision but performance gains are limited by the compression rate. High compression rate often leads to poor discriminative representations. To this end, this paper aims to enhance the discriminative power of feature representations from a new feature-learning perspective. Specifically, we attempt to learn more disciminative representations with contrastive instances for UAV tracking in a simple yet effective manner, which not only requires no manual annotations but also allows for developing and deploying a lightweight model. We are the first to explore contrastive learning for UAV tracking. Extensive experiments on four UAV benchmarks, including UAV123@10fps, DTB70, UAVDT and VisDrone2018, show that the proposed DRCI tracker significantly outperforms state-of-the-art UAV tracking methods.
\end{abstract}


\section{Introduction}
UAV tracking aims to infer and predict the location and scale of arbitrary objects in consecutive aerial image frames and has a broad range of potential applications in navigation, agriculture, transportation, disaster response, and public safety~\cite{li2020autotrack,cao2021hift,Wang2022RankBasedFP,Wu2022FisherPF,articlewxc}. Compared with general object tracking, UAV tracking is challenging due to motion blur, severe occlusion, extreme viewing angle, and scale changes, making it difficult to achieve high precision. In addition,  limited computing resources, low power requirements, battery capacity limitations, and the maximum load of UAVs also pose a considerable challenge to tracking efficiency~\cite{Wang2022RankBasedFP,Wu2022FisherPF,li2021learning}.

Maintaining high efficiency and high precision are two fundamental challenges in UAV tracking. Discriminative correlation filters (DCF)-based trackers dominate in this field because of their high efficiency on a single CPU. However, their precisions are not comparable to most cutting-edge deep learning (DL)-based trackers~\cite{li2020autotrack, huang2019learning, Zhang2022TrackingSA,Zhang2023TsfmoAB}. DL-based trackers are well known for their high precision, but they usually rely on complex architecture, leading to low efficiency. To combat efficiency drop, some lightweight DL-based trackers have recently been proposed  for UAV tracking~\cite{Wang2022RankBasedFP,Wu2022FisherPF,Liu2022GlobalFP,Zhong2022EfficiencyAP}, which mainly utilize model compressing techniques such as filter pruning to boost efficiency while maintaining high precision. Unfortunately, the filter pruning methods utilized by these works such as rank-based filter pruning \cite{Wang2022RankBasedFP} and Fisher pruning \cite{Wu2022FisherPF}, though simple, the achieved tracking precision and efficiency are very limited and far from satisfactory. The performance limitation is because the high compression rates of these methods are prone to produce inferior discriminative representations. To this end, in this paper, we explore dealing with low performance in UAV tracking from a new feature-learning perspective to enhance the discriminative power of feature representations.

Contrastive learning is a discriminative approach that aims to learn an embedding space where similar sample pairs~(aka positive pairs) stay close to each other and dissimilar ones~(aka negative pairs) are far apart. It has been successfully used in many vision tasks such as image classification~\cite{chen2020simple}, image-to-image translation~\cite{Park2020ContrastiveLF}, text-to-image generation~\cite{Zhang2021CrossModalCL}, and natural language understanding~\cite{Li2022PairLevelSC}. It is worth nothing that contrastive learning has also been applied to single object tracking~\cite{Wu2021ProgressiveUL,pi2022hierarchical} and multiple object tracking~\cite{Pang2021QuasiDenseSL,Yu2022TowardsDR}. However, these methods usually require collecting additional annotations for positive pairs which is expensive and time-consuming~\cite{pi2022hierarchical}. Or contrastive learning of these methods is intertwined with heavy and complicated tracking frameworks~\cite{Pang2021QuasiDenseSL,Wu2021ProgressiveUL,Yu2022TowardsDR}, making it impossible to transfer the learning mechanism to UAV tracking. In this paper, we attempt to utilize contrastive learning  for UAV tracking in a simple yet effective manner, which not only requires no manual annotations but also allows for developing and deploying a lightweight model.

Specifically, we use intra- and inter-video templates of targets as our contrastive instances to facilitate discriminative representation learning for UAV tracking. Unlike classic contrastive learning~\cite{chen2020simple} where positive pairs are constructed from image augmentation, we construct positive pairs from a video. To avoid selecting hard positive samples~(e.g., occluded target), we empirically randomly select 2 frames from the video to construct positive sample pairs as we observe most of the positive sample pairs are of good quality. As a result, the proposed tracker learns discriminative representations with contrastive instances~(DRCI), which achieves state-of-the-art efficiency and precision compared with existing CPU-based and lightweight DL-based trackers in UAV tracking. In the inference stage, there is no additional computation burden when applying our DRCI. 

To sum up, this paper makes the following contributions:

\begin{itemize}
    \item We make the first attempt to explore contrastive learning for UAV tracking, a new feature-learning perspective to obtain lightweight DL-based trackers with better tracking precision and efficiency.
	\item We propose the DRCI tracker that learns discriminative representations with contrastive instances, achieving a remarkable balance between tracking efficiency and precision. 
	\item We demonstrate the proposed method on four public UAV benchmarks. Experimental results show that the proposed DRCI tracker achieves state-of-the-art performance.
\end{itemize}

\section{Related Work}

\subsection{UAV Tracking Methods}

Modern trackers can be roughly divided into two categories: DCF-based trackers and DL-based trackers. The former dominates in UAV tracking with its more favorable efficiency. DCF-based trackers start with a minimum output sum of squared error (MOSSE) filter. Since then, DCF-based trackers have made great progress in many variants \cite{li2021learning}, including state-of-the-art UAV tracking methods \cite{li2021learning,Li2021LearningRC,li2020autotrack,huang2019learning,li2020asymmetric,Li2021EquivalenceOC}. Despite their relatively higher efficiency, they are difficult to maintain robustness under challenging conditions due to the poor representation ability of handcrafted features.

Thanks to the powerful feature representation ability, deep learning has proven to be very successful in visual tracking in recent years. To substantially improve tracking precision and robustness, some DL-based trackers have recently been developed for UAV tracking. For instance, Cao et al. \cite{cao2021hift} proposed a hierarchical feature transformer to enable interactive fusion of spatial (shallow layers) and semantics cues (deep layers) for UAV tracking. Fu et al. \cite{Fu2021SiameseAP} proposed a two-stage Siamese network-based method in which high-quality anchor proposals are generated in stage 1 and then refined in stage 2. Cao et al. \cite{Cao2022TCTrackTC} proposed a comprehensive framework to fully exploit temporal contexts with an adapative temporal transformer for aerial tracking. However, the efficiency of these methods is still much lower than most DCF-based trackers. To further improve the efficiency of DL-based trackers for UAV tracking, model compression techniques have been recently utilized to reduce model size \cite{Wang2022RankBasedFP,Wu2022FisherPF}. Unfortunately, the model compression methods used by these works, although simple, still cannot achieve satisfying tracking precision at higher compression rates. In contrast, in this paper, we explore dealing with low performance in UAV tracking from a new feature-learning perspective~(i.e., contrastive learning) to enhance the discriminative power of feature representations.

\subsection{Contrastive Learning}
Contrastive learning aims at learning representations by contrasting between similar and dissimilar samples. Specifically, it attempts to bring similar samples closer together in the representation space while pushing dissimilar ones apart. It has received a great deal of attention because of its outstanding performance in the field of self-supervised learning \cite{chen2020simple, Park2020ContrastiveLF, Zhang2021CrossModalCL, Li2022PairLevelSC}. Although contrastive learning has been deployed in many fields, until recently it was applied to multiple object tracking \cite{Pang2021QuasiDenseSL,Yu2022TowardsDR} and single object tracking \cite{Wu2021ProgressiveUL,pi2022hierarchical}. For instance, Pang et al. \cite{Pang2021QuasiDenseSL} presented a quai-dense similarity learning that densely samples hundreds of region proposals on a pair of images for contrastive learning to exploit most informative regions on images. Yu et al. \cite{Yu2022TowardsDR} proposed a trajectory-level contrastive loss to exploit the inter-frame information contained in the entire trajectory of a certain target. Wu et al. \cite{Wu2021ProgressiveUL} proposed a progressive unsupervised learning (PUL) framework, which is the first discrimination model that learn to effectively distinguish objects from backgrounds in a contrastive learning manner. Pi et al. \cite{pi2022hierarchical} developed instance-aware and category-aware modules to exploit different semantic levels with contrastive learning to produce robust feature embeddings. However, these methods usually require collecting additional annotations for positive pairs which is expensive and time-consuming~\cite{pi2022hierarchical}. Or contrastive learning of these methods is intertwined with heavy and complicated tracking frameworks~\cite{Pang2021QuasiDenseSL,Wu2021ProgressiveUL,Yu2022TowardsDR}, making it impossible to transfer the learning mechanism to UAV tracking. In this paper, we attempt to leverage contrastive learning in a simple yet effective manner to achieve more discriminative feature representations to improve both precision and efficiency of lightweight DL-based trackers for UAV tracking. 

\section{Learning Discriminative Representation with Contrastive Instances}

\begin{figure*}[h]
	\centering
	\includegraphics[width=0.98\textwidth,height=0.32\textwidth]{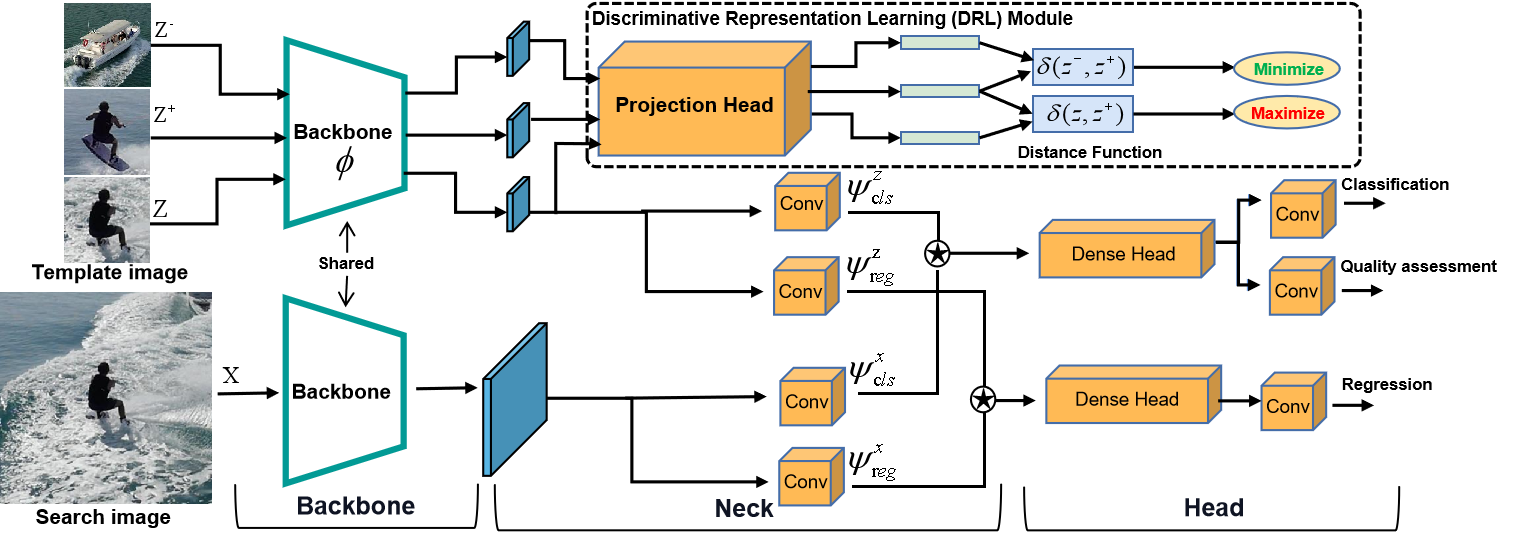}
	\caption{An illustration of the proposed DRCI method. Note that $\psi_{cls}^{\cdot}$ and $\psi_{reg}^{\cdot}$ denote the task-specific convolutional layers for classification and regression, respectively. The template $Z$ is taken as an anchor in our contrastive learning while $Z^+$ and $Z^-$ are positive and negative samples, respectively. } \label{P-saimfc++_overview}
\end{figure*}

\subsection{DRCI Overview}
As illustrated in Fig. \ref{P-saimfc++_overview}, the proposed DRCI consists of a backbone, a neck, a head network and a discriminative representation learning (DRL) module. Specifically, the backbone network $\phi(\cdot)$ is a Siamese network, shared by the template branch and the search branch, which take template image $Z$ and search image $X$ as input, respectively. The neck contains four convolutional layers to adjust feature sizes. The head consists of two dense head branches followed by three convolutional layers to produce outputs for classification, quality assessment, and regression tasks. Backbone features from two branches are adjusted at the neck and then coupled with cross-correlation before they are finally fed into the classification and regression heads. The coupling features are formulated by:
\begin{equation}
\small
    f_l(Z,X) = E_2(\psi_l^z(\phi(Z))) \star E_2(\psi_l^x(\phi(X))), l\in \{cls, reg\},
\end{equation}
where $\star$ denotes the cross-correlation operation, $E_2$ represents the encoder for identity-related feature embedding. $\psi_{cls}^x(\cdot)$ and $\psi_{reg}^x(\cdot)$ denote the task-specific layer for classification and regression, respectively, with the same output size. $\psi_{cls}^z(\cdot)$ and $\psi_{reg}^z(\cdot)$ have the similar meaning. In the training stage, we use a DRL module to enhance the discriminative power of feature representations for UAV tracking. In the inference stage, the DRL module is removed, so there is no additional computation burden when applying our DRCI. 

\subsection{Discriminative Representation Learning (DRL)}
The DRL module utilizes a project head $Proj(\cdot)$ to project the backbone features into an embedding space that the similarity of the backbone features, hopefully, can be well evaluated by a relatively simple distance function. For simpilicty, we instantiate the projection head as fully connected layer followed by a ReLU activation, similar to SimCLR \cite{chen2020simple}. A more refined design of the projection head could lead to further performance improvements, which we leave for future research. To obtain instance samples for contrastive learning, we first randomly sample a minibatch of N frame pairs from N different sequences. We then crop the target templates from each frame, yielding N positive pairs and ($C_N^2-N$) negative constrative pairs. Denote these contrastive template samples as $\{Z_i\}_{i=1}^{2N}$, let $I\equiv\{1,...,2N\}$ and $j(i)$ be the index of the other sample originating from the same target, i.e., $Z_i$ and $Z_{j(i)}$ make a positive pair, denoted by $Z_i\leftrightarrow  Z_{j(i)}$. We adopt the supervised contrastive loss proposed in \cite{khosla2020supervised} for our discriminative representation learning, except that the negative sample pairs are pseudo or not ground truth, which takes the following form,
\begin{equation}\label{Eq_JSDMI}
\small
\begin{split}
L_{DRL}=\sum_{i\in I}\frac{1}{\left | P(i)\right |}\sum_{p\in P(i)}log\frac{exp(z_i\cdot z_p/\tau)}{\sum_{a\in A(i)}exp(z_i\cdot z_a/\tau)},
\end{split}
\end{equation}
where $z_i=Proj(\phi(Z_i))$, $\cdot$ denotes the inner product, $\tau\in \mathbb{R}^+$ is a scalar temperature parameter, $A(i)=I\setminus\{i\}$, $P(i)=\{p\in A(i): Z_p \leftrightarrow  Z_i \}$ is the set of indices of all positive samples in the minibatch of $i$ except itself, and $\left | P(i)\right |$ denotes the cardinality of $P(i)$. The DRL loss tries to increase the similarity between feature representations of the targets in the same sequence, while suppressing that of different sequences. 

\subsection{Classification, Regression and Quality Assessment Losses}
The classification branch predicts the category for each location and the regression branch calculates the target bounding box for that location. The outputs of two branches are represented as $O_{h\times w\times 2}^{cls}$ and $O_{h\times w\times 4}^{reg}$, respectively, and $w$ and $h$ denote the width and height. Specifically, $O_{h\times w\times 2}^{cls}(i,j,:)$ is a 2D vector, representing the foreground and background scores at position $(i,j)$. $O_{h\times w\times 4}^{reg}(i,j,:)$ is a 4D vector, representing the distances from the corresponding position to the four sides of the bounding box. At the same time, the quality assessment branch, with output being $O_{h\times w\times 1}^{qs}$, is in parallel with the classification branch to assess classification quality, which is finally used to reweight the classification score. Following P-SiamFC++ \cite{Wang2022RankBasedFP}, the losses for learning these tasks is as follows:
\begin{equation}\label{EQ_RRCF}
	\small
	\begin{split}
		L_{CRQ}= \frac{1}{N_{pos}}\sum_{z}( L_{cls}(p_z, p_z^*) + \lambda_1 I_{\{p_z^*>0\}}L_{reg}(t_z,t_z^*)+\\ \lambda_2 I_{\{p_z^*>0\}}L_{qs}(q_z,q_z^*))
	\end{split}
\end{equation}
\begin{figure*}[t]
	\centering
	\subfigure{
		\begin{minipage}[t]{0.23\textwidth}
			\includegraphics[width=1\textwidth,height=0.68\textwidth]{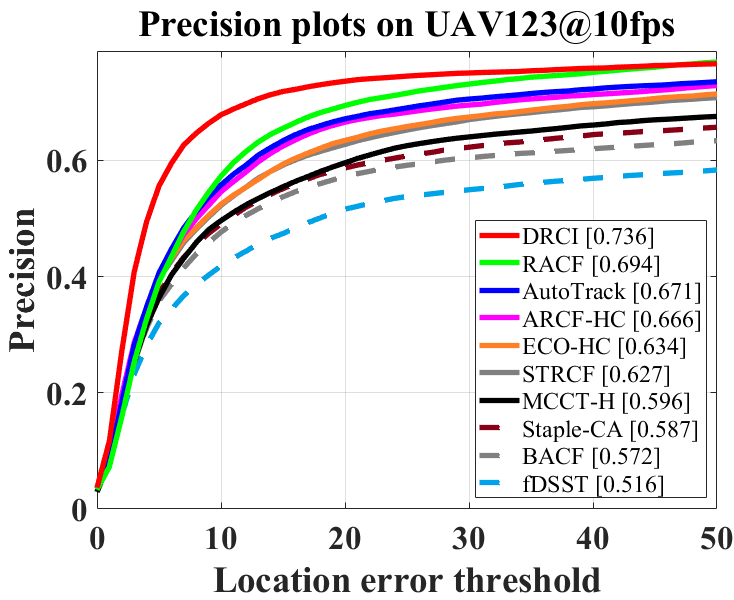}\hspace{0in}
			\includegraphics[width=1\textwidth,height=0.68\textwidth]{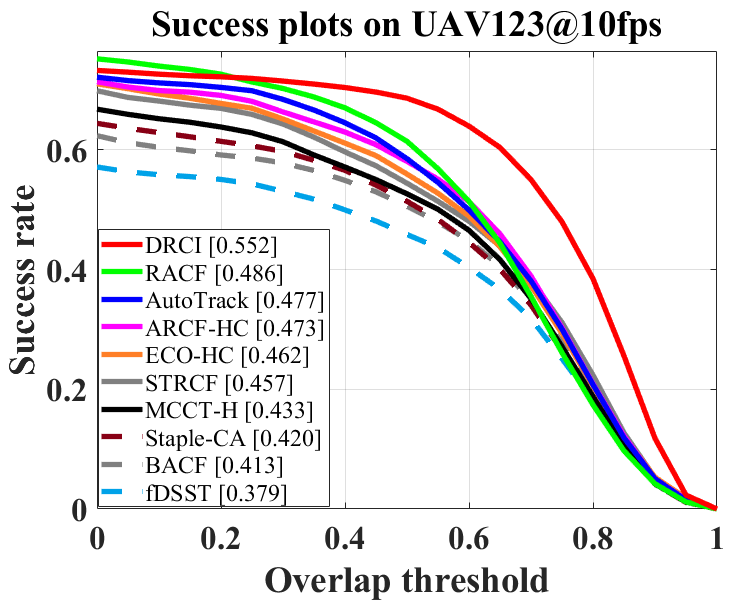}
	\end{minipage}}
	\subfigure{
		\begin{minipage}[t]{0.23\textwidth}
			\includegraphics[width=1\textwidth,height=0.68\textwidth]{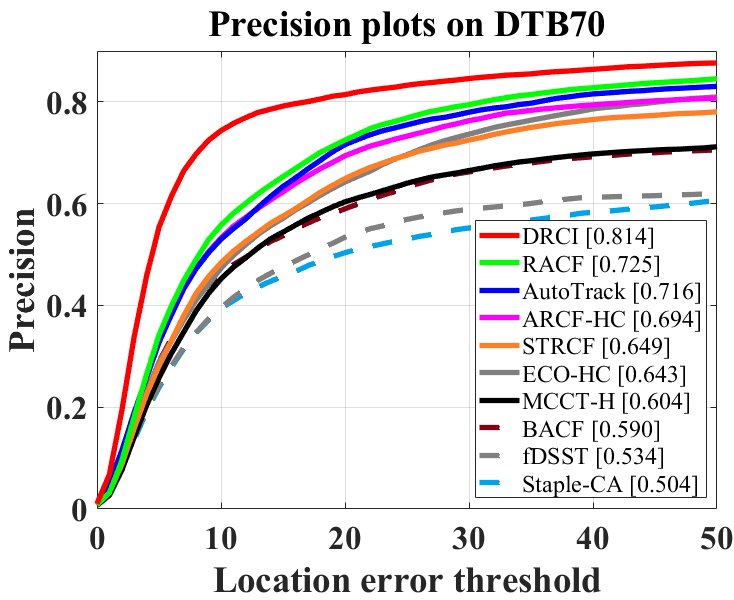}\hspace{0in}
			\includegraphics[width=1\textwidth,height=0.68\textwidth]{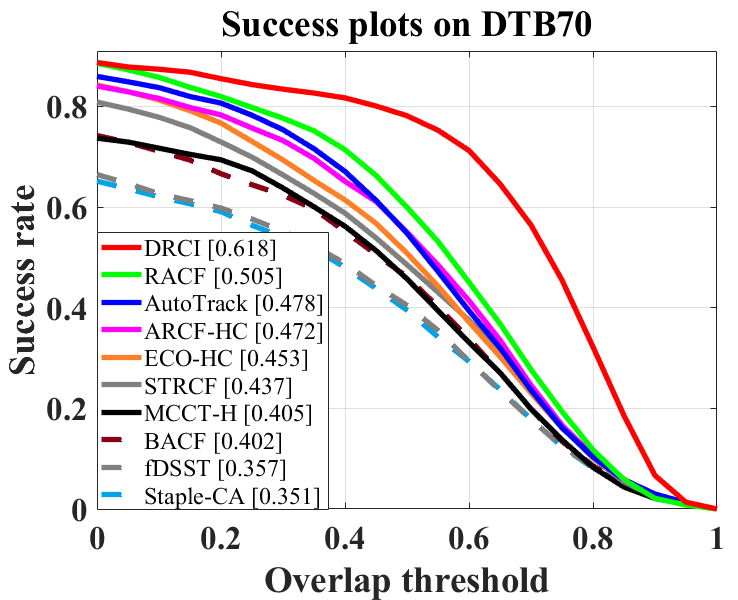}
	\end{minipage}}
	\subfigure{
		\begin{minipage}[t]{0.23\textwidth}
			\includegraphics[width=1\textwidth,height=0.68\textwidth]{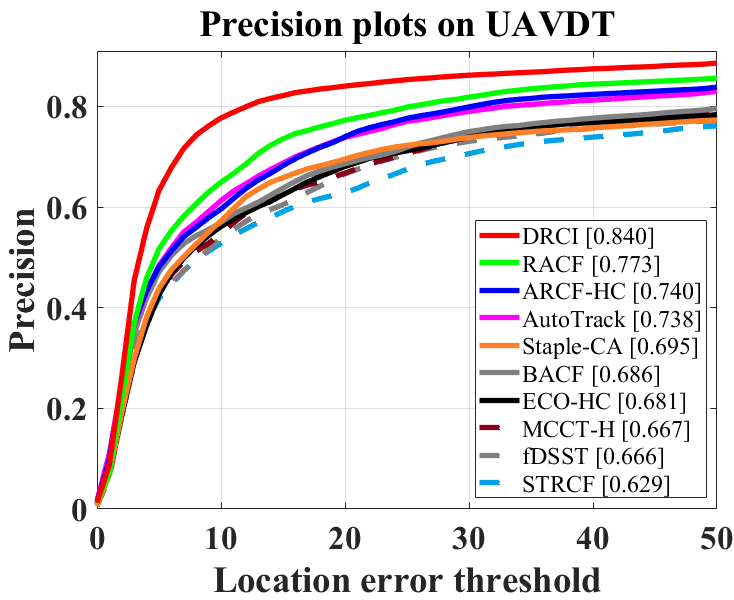}\hspace{0in}
			\includegraphics[width=1\textwidth,height=0.68\textwidth]{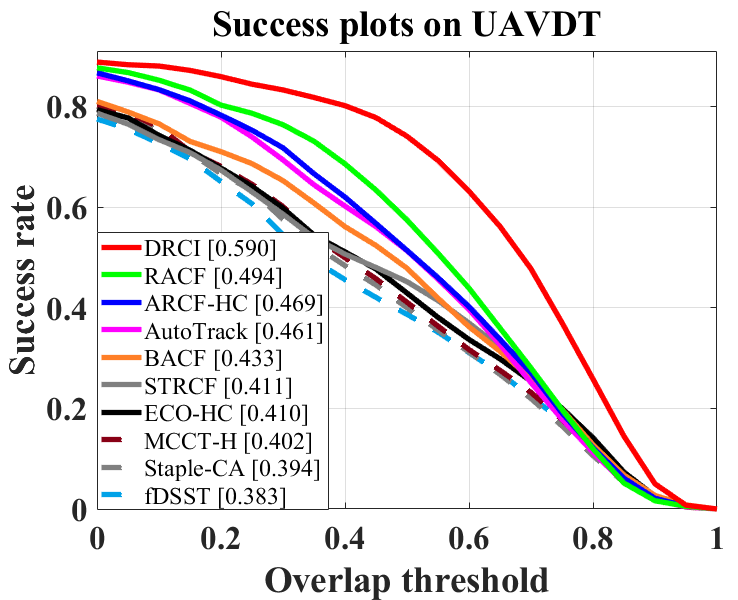}
	\end{minipage}}
	\subfigure{
		\begin{minipage}[t]{0.23\textwidth}
			\includegraphics[width=1\textwidth,height=0.68\textwidth]{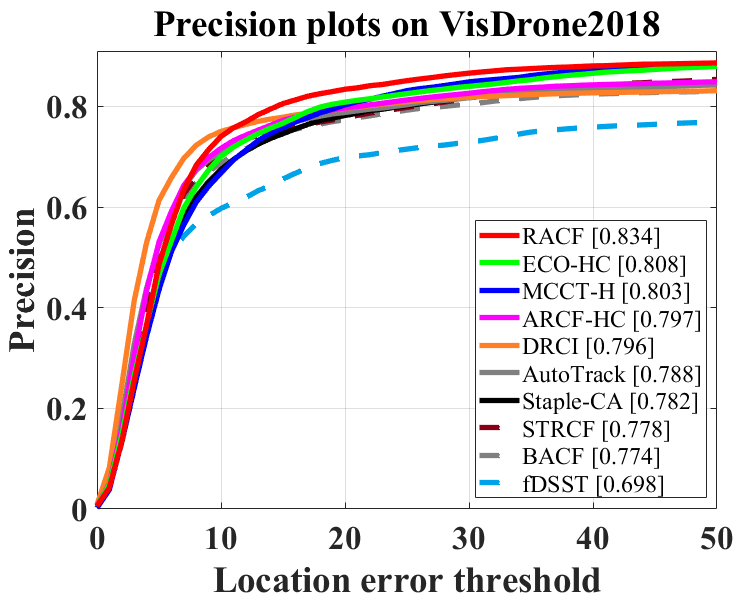}\hspace{0in}
			\includegraphics[width=1\textwidth,height=0.68\textwidth]{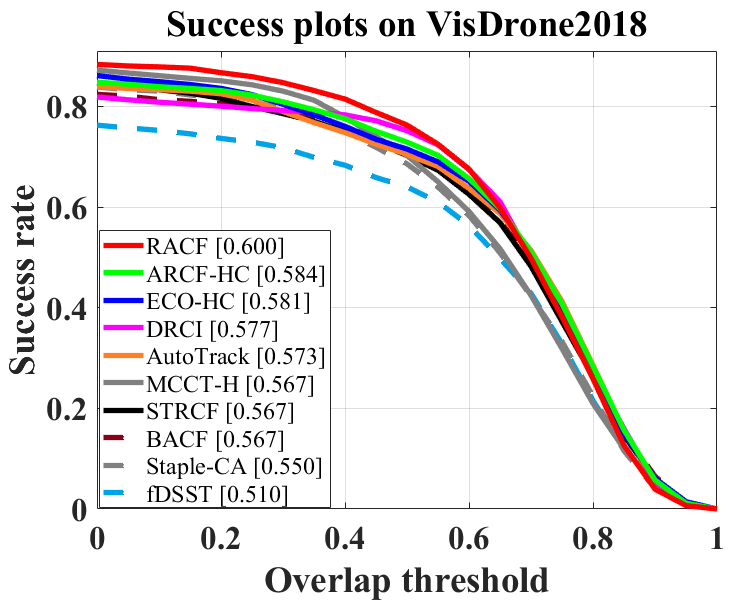}
	\end{minipage}}
	\caption{Overall performance of hand-crafted based trackers on datasets, from left to right, UAV123@10fps, DTB70, UAVDT and VisDrone2018. Precision and success rate for one-pass evaluation (OPE) are used for evaluation. The precision at 20 pixels and area under curve (AUC) are used for ranking, marked in the precision plots and success plots, respectively.}
	\label{fig_overall_p_s_plots}
 \end{figure*}
\begin{table*}[h]
\centering
\caption{Average precision and speed (FPS) comparision between DRCI and hand-crafted based trackers on UAV123@10fps, DTB70, UAVDT and VisDrone2018.  All the reported FPSs are evaluated on a single CPU. {\color[HTML]{FE0000}Red}, {\color[HTML]{3531FF}blue} and {\color[HTML]{009901}green} respectively indicate the first, second and third place.}
\label{tab:precision_FPS}
\resizebox{7.0in}{0.3in}{
\begin{tabular}{cccccccccc} 
\toprule
                   & KCF\cite{2015High}& fDSST\cite{danelljan2016adaptive}& BACF\cite{2017Learning}& ECO-HC\cite{danelljan2017eco}&  STRCF\cite{li2018learning} & ARCF-HC\cite{huang2019learning}  & AutoTrack\cite{li2020autotrack}  		   &RACF\cite{li2021learning}                            & \textbf{DRCI (Ours)}            \\ 
\hline\hline
\textbf{Precision} & 53.3                            & 60.4                             & 64.2 & 68.8                                      & 67.1  & 71.9    & \textbf{\textcolor[rgb]{0,0.502,0}{72.3}} & \textbf{\textcolor{blue}{75.7}} & \textbf{\textcolor{red}{79.7}}  \\
\textbf{FPS (CPU)} & \textbf{\textcolor{red}{622.5}} & \textbf{\textcolor{blue}{193.4}} & 54.2 & \textcolor[rgb]{0,0.502,0}{\textbf{84.5}} & 28.4  & 34.2    & 58.7                                      & 35.7                            & 58.9                            \\
\hline
\end{tabular}
}
\end{table*}
where $L_{cls}$, $L_{reg}$ and $L_{qs}$ denote the focal loss, the IoU loss and the binary cross entropy loss for classification, regression and quality assessment, respectively. $z$ represents a coordinate on a feature map, $p_z$ is a prediction while $p_z^*$ is the corresponding target label, $I_{\{\cdot\}}$ is the indicator function, $N_{pos}=\sum_{z}I_{\{p_z^*>0\}}$. $\lambda_1$ and $\lambda_2$ are weight terms to balance the losses. Note that $p_z^*$ is assigned 1 if $z$ is considered a positive sample, otherwise 0 if it is considered a negative sample. 

Taken together, the overall loss for training our DRCI is:
\begin{equation}\label{EQ_totalloss}
\small
	\begin{split}
		L = L_{CRQ} + \rho L_{DRL},
	\end{split}
\end{equation}
where $\rho$ is a constant coefficient to balance $L_{CRQ}$ and $L_{DRL}$.

\section{Experiments}

\begin{table*}[]
	\caption{Precision and speed (FPS) comparison between DRCI and
		deep-based trackers on UAVDT \cite{du2018the}. All the reported FPSs are evaluated on a single GPU. {\color[HTML]{FE0000}Red}, {\color[HTML]{3531FF}blue} and {\color[HTML]{009901}green} indicate the first, second and third place. }
	\label{tab:precision_FPS_deep}
 	\resizebox{7.0in}{0.32in}{
		\begin{tabular}{cccccccccc}
			\toprule
  &\begin{tabular}[c]{@{}c@{}}SiamGAT\cite{guo2021graph}\end{tabular} 
  &  \begin{tabular}[c]{@{}c@{}}HiFT\cite{cao2021hift}\end{tabular}
  & \begin{tabular}[c]{@{}c@{}}AutoMatch\cite{zhang2021learn}\end{tabular} 

         &\begin{tabular}[c]{@{}c@{}}TCTrack\cite{Cao2022TCTrackTC}\end{tabular}
         &\begin{tabular}[c]{@{}c@{}}F-SiamFC++\cite{Wu2022FisherPF}\end{tabular}
         &\begin{tabular}[c]{@{}c@{}}P-SiamFC++\cite{Wang2022RankBasedFP}\end{tabular}
            
         &\begin{tabular}[c]{@{}c@{}}SLT-TransT\cite{Kim2022TowardsST}\end{tabular} 
         &\begin{tabular}[c]{@{}c@{}}SparseTT\cite{2022SparseTT}\end{tabular}
     & \begin{tabular}[c]{@{}c@{}}\textbf{DRCI (Ours)} \end{tabular} 
          \\ \hline\hline
          
	\textbf{Precision}  &76.4 &65.2    &73.8      &69.6 & 79.4                & 80.7                   & {\color[HTML]{009901} \textbf{82.9}}  & {\color[HTML]{3531FF} \textbf{82.8}}    
 & {\color[HTML]{FE0000} \textbf{84.0}}  
 
    \\\textbf{FPS (GPU)}       
 
			 &71.0 &137.3  &43.1  &125.7  &{\color[HTML]{3531FF} \textbf{266.2}}  &{\color[HTML]{009901} \textbf{258.8}}  
    &29.9 &45.1 & {\color[HTML]{FE0000} \textbf{298.3}}    \\ \hline
		\end{tabular}
	}
\end{table*}

We conduct our experiments on four challenging UAV benchmarks, i.e., UAV123@10fps \cite{2016A}, DTB70 \cite{li2017visual}, UAVDT \cite{du2018the} and VisDrone2018 \cite{wen2018visdrone}. All evaluation experiments are conducted on a PC equipped with i9-10850K processor (3.6GHz), 16GB RAM and an NVIDIA TitanX GPU. The backbone, neck, and head architectures are inherited from F-SiamfC++ but with block-wise pruning ratios of 0.7, 0.5 and 0.3, respectively. The temperature parameter $\tau$ is set to 0.5. The default setting of $\rho$ is 0.1 and other parameters such as $\lambda_1$ and $\lambda_2$ for training and inference follow P-SiamFC++. Code wiil be  available on: \url{https://github.com/P-SiamFCpp/DRCI}. 

\subsection{Comparison with CPU-based Trackers}

Eight state-of-the-art trackers based on hand-crafted features for comparison are: KCF \cite{2015High}, fDSST \cite{danelljan2016adaptive},  BACF \cite{2017Learning}, ECO-HC \cite{danelljan2017eco}, STRCF \cite{li2018learning}, ARCF-HC \cite{huang2019learning}, AutoTrack  \cite{li2020autotrack}, RACF \cite{li2021learning}. 

The overall performance of DRCI with the competing trackers on the four benchmarks is shown in Fig. \ref{fig_overall_p_s_plots}. It can be seen that DRCI outperforms all other trackers on all benchmarks except for the VisDrone2018.
Specifically, on UAV123@10fps, DTB70 and UAVDT, DRCI significantly outperforms the second tracker RACF in terms of precision and AUC, with gains of (4.2\%, 6.6\%), (8.9\%, 11.3\%) and (6.7\%, 9.6\%), respectively. On VisDrone2018, our DRCI is inferior to the first tracker RACF in precision and AUC, the gaps are 3.8\% and 2.3\%, respectively. The reason is that the parameters of RACF is dataset specific, while our DRCI is not. DRCI is also slightly better than ECO-HC, MCCT-H, and ARCF-HC in precision with a max gap being 1.1\%, and surpassed by ARCF-HC and ECO-HC in AUC with a max gap being 0.7\%. 
In terms of speed, we use the average FPS over the aforementioned four benchmarks on CPU as a tracking metric. Table \ref{tab:precision_FPS} illustrates the average precision and FPS produced by different trackers. It can be seen that DRCI outperforms all competing trackers in precision, and is the best real-time tracker (speed of $>$30FPS) on CPU. Specifically, DRCI achieves 79.7\% in precision at a speed of 58.9 FPS.

\begin{table*}
\centering
\caption{Comparison of model size (parameters), precision and tracking speed between the proposed DRCI and the baseline method P-SiamFC++ on four UAV benchmarks. PRC is short for precision. Note that only the precision on CPU is shown here since the difference of precision on CPU and GPU is very small.}
\label{tab:compareWbaseline}
\resizebox{7.0in}{0.515in}{
\begin{tabular}{ccccccccccccccccc} 
\toprule
\multirow{3}{*}{Methods} & \multirow{3}{*}{Parameters} & \multicolumn{3}{c}{UAV123@10fps}                      & \multicolumn{3}{c}{DTB70}                             & \multicolumn{3}{c}{UAVDT}                             & \multicolumn{3}{c}{VisDrone2018}                      & \multicolumn{3}{c}{Avg.}                               \\ 
\cmidrule(l){3-17}
                         &                             & \multirow{2}{*}{PRC} & \multicolumn{2}{c}{FPS}        & \multirow{2}{*}{PRC} & \multicolumn{2}{c}{FPS}        & \multirow{2}{*}{PRC} & \multicolumn{2}{c}{FPS}        & \multirow{2}{*}{PRC} & \multicolumn{2}{c}{FPS}        & \multirow{2}{*}{PRC} & \multicolumn{2}{c}{FPS}         \\ 
\cmidrule(l){4-5}\cmidrule(l){7-8}\cmidrule(l){10-11}\cmidrule(l){13-14}\cmidrule(l){16-17}
                         &                             &                      & CPU           & GPU            &                      & CPU           & GPU            &                      & CPU           & GPU            &                      & CPU           & GPU            &                      & CPU           & GPU             \\ 
\hline\hline
P-SiamF++~\cite{Wang2022RankBasedFP}              & 7.49M                       & 73.1                 & 45.1          & 236.4          & 80.3                 & 45.6          & 238.2          & 80.7                 & 48.8          & 258.8          & \textbf{80.9}        & 45.0          & 230.5          & 78.8                 & 46.1          & 241.0           \\
\textbf{DRCI (Ours)}              & \textbf{5.05M}                       & \textbf{73.6}        & \textbf{59.2} & \textbf{300.7} & \textbf{81.4}        & \textbf{60.1} & \textbf{297.7} & \textbf{84.0}        & \textbf{59.4} & \textbf{298.3} & 79.6                 & \textbf{57.0} & \textbf{284.6} & \textbf{79.7}        & \textbf{58.9} & \textbf{295.3}  \\
\bottomrule
\end{tabular}
}
\end{table*}

\begin{figure}[h]
	\centering
	\includegraphics[width=0.49\textwidth,height=0.35\textwidth]{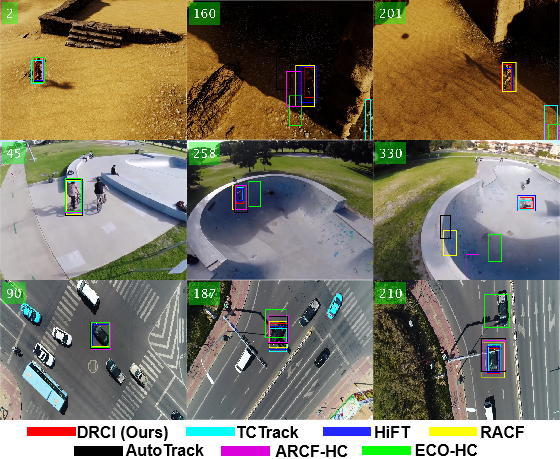}
	\caption{Qualitative evaluation on 3 sequences from, respectively, UAV123@10fps, DTB70 and UAVDT (i.e. person1\_s, BMX4 and S0309). The results of different methods are represented by different colors.} \label{fig:visual_examples}
\end{figure}

\subsection{Comparison with DL-based Trackers}

The proposed DCRI is also compared with eight state-of-the-art DL-based trackers on the UAVDT dataset \cite{wen2018visdrone}, including SiamGAT \cite{guo2021graph}, HiFT \cite{cao2021hift}, AutoMatch \cite{zhang2021learn}, SLT-SiamRPN++ \cite{Kim2022TowardsST}, SparseTT \cite{2022SparseTT},
TCTrack \cite{Cao2022TCTrackTC}, F-SiamFC++ \cite{Wu2022FisherPF}, P-SiamFC++ \cite{Wang2022RankBasedFP}. 

The FPSs and the precisions on UAVDT are shown in Table \ref{tab:precision_FPS_deep}. As can be seen, the precision and the GPU speed of our DRCI outperform that of the competing DL-based trackers, surpassing the second tracker SparseTT \cite{2022SparseTT} by 1.2\% in precision, and its GPU speed is more than 6 times faster than the second tracker SparseTT \cite{2022SparseTT}. This not only verifies that the proposed method can obtain a lightweight DL-based tracker with better tracking precision and efficiency, but also supports our solution to address the low performance in UAV tracking from a new feature-learning perspective, which indeed enhances the discriminative power of feature representations.

\subsection{Qualitative Comparison with SOTA Trackers}
We show some qualitative tracking results of our method and six state-of-the-art trackers in Fig. \ref{fig:visual_examples}. As can be seen, only our tracker DRCI successfully track the targets in all three challenging examples, where the objects are experiencing illumination change (i.e., persion1\_s and BMX4) or pose variations (i.e., BMX4 and S0309). Our method performs much better and is more visually pleasing in these cases, further supporting the effectiveness of the proposed method of learning discriminative representation using contrastive instances for UAV tracking.

\subsection{Ablation Study}
\indent\textbf{Effect of Discriminative Representation Learning (DRL):} 
We compare the proposed DRCI with the baseline P-SiamFC++ on all four UAV benchmarks in terms of model size, precision and tracking speed to understand its effectiveness. Their comparisons are shown in Table \ref{tab:compareWbaseline}. As can be seen, the model size of DRCI is reduced to 67.4\% ($\approx$5.05/7.49) of the original. Both CPU and GPU speed have been increased. Specifically, on average, the CPU speed increased from 46.1 FPS to 58.9 FPS while the GPU speed increased from 241.0 FPS to 295.3 FPS. Although DRCI is slightly inferior to the baseline on VisDrone2018 in precision by 1.3\%, the improvement on DTB70 and UAVDT is significant, specifically, with gains of 1.1\% and 3.3\%, respectively. These results justify that the effectiveness of using DRL~(a new feature-learning perspective) to assist UAV tracking by improving both efficiency and precision.

\begin{table}
\centering
\caption{Illustration of how the precision of DRCI on the four benchmarks varies with the weight~(i.e., $\rho$.) of the loss of discriminative representation learning.  }
\label{tab:Impact_of_ratios}
\resizebox{3.4in}{0.40in}{
\begin{tabular}{ccccccccccc} 
\toprule
$\rho$          & 0.0                             & 0.1                             & 0.2                             & 0.3                                       & 0.4  & 0.5                                       & 0.6                                       & 0.7  & 0.8                             & 0.9   \\ \hline
\midrule
DTB70        & \textcolor{blue}{\textbf{80.5}} & \textcolor{red}{\textbf{81.5}}  & 80.1                            & 78.9                                      & 79.0 & \textbf{\textcolor[rgb]{0,0.502,0}{80.4}} & 78.6                                      & 78.1 & 78.9                            & 77.9  \\
UAVDT        & 76.2                            & \textbf{\textcolor{red}{84.0}}  & \textcolor{blue}{\textbf{82.7}} & \textcolor[rgb]{0,0.502,0}{\textbf{81.9}} & 78.9 & 80.8                                      & 81.8                                      & 78.9 & 76.5                            & 79.5  \\
UAV123@10fps & \textcolor{red}{\textbf{72.8}}  & \textbf{\textcolor{blue}{72.1}} & 69.9                            & 70.0                                      & 69.4 & 70.8                                      & \textcolor[rgb]{0,0.502,0}{\textbf{71.2}} & 70.7 & 69.3                            & 69.5  \\
VisDrone2018 & 72.5                            & \textcolor{red}{\textbf{79.6}}  & 76.9                            & \textcolor[rgb]{0,0.502,0}{\textbf{77.4}} & 76.4 & 76.0                                      & 76.0                                      & 74.5 & \textcolor{blue}{\textbf{77.5}} & 74.5  \\
\bottomrule
\end{tabular}
}
\end{table}

\noindent\textbf{Impact of loss $L_{DRL}$:} To see how the DRL loss affects the precision of DRCI, we train DRCI with different DRL loss weights and evaluate on four benchmarks. The weight $\rho$ (refer to Eq. \ref{EQ_totalloss}) ranges from 0.0 to 1.0 in step of 0.1. Table \ref{tab:Impact_of_ratios} shows the precision of DRCI with different $\rho$ on four benchmarks. Note that $\rho=0.0$ represents the baseline tracker P-SiamFC++. It can be seen that when $\rho$ is 0.1, DRCI achieves the best precision on four benchmarks except UAV123@10fps. Remarkably, significant improvements can be seen on UAVDT and VisDrone2018 with $\rho>0.0$, namely imposing the proposed DRL loss, although the precision fluctuates on DTB70 and UAV123@10fps. Overall, the best precisions occur when $\rho$ is about 0.1. This result suggests that appropriately imposing the proposed DRL loss can help improve the precision of the baseline tracker, justifying the effectiveness of the proposed DRCL.

\section{Conclusion}
In this work, we are the first to explore learning discriminative representation with contrastive instances for UAV tracking, which not only requires no manual annotations but also allows for developing and deploying a lightweight model. The proposed DRCI is able to learn more effective and more compact representations, and demonstrates state-of-the-art performance on four UAV benchmarks in terms of efficiency as well as tracking precision. We believe our work will draw more attention to developing more effective and more efficient lightweighted DL-based trackers for UAV tracking.

\section*{Acknowledgment}
Thanks to the supports by Guangxi Key Laboratory of Embedded Technology and Intelligent System, Research Institute of Trustworthy Autonomous Systems, the National Natural Science Foundation of China (No. 62176170, 62066042, 61971005), the Science and Technology Department of Tibet (No. XZ202102YD0018C), the Sichuan Province Key Research and Development Project (No. 2020YJ0282), and the Guangxi Science and Technology Base and Talent Special Project (No. 2021AC9330).

\small
\bibliographystyle{IEEEbib}
\bibliography{icme}

\begin{thebibliography}{10}

\bibitem{li2020autotrack}
{Li Y. and et al.},
\newblock ``Autotrack: Towards high-performance visual tracking for uav with
  automatic spatio-temporal regularization,''
\newblock in {\em CVPR,2020}, pp. 11923--11932.

\bibitem{cao2021hift}
Cao Z. and et~al.,
\newblock ``Hift: Hierarchical feature transformer for aerial tracking,''
\newblock in {\em ICCV}, 2021, pp. 15457--15466.

\bibitem{Wang2022RankBasedFP}
Wang X. and et~al.,
\newblock ``Rank-based filter pruning for real-time uav tracking,''
\newblock {\em ICME}, pp. 01--06, 2022.

\bibitem{Wu2022FisherPF}
Wu~W. and et~al.,
\newblock ``Fisher pruning for real-time uav tracking,''
\newblock {\em IJCNN}, pp. 1--7, 2022.

\bibitem{articlewxc}
Wang X. and et~al.,
\newblock ``Exploiting rank-based filter pruning for real-time uav tracking,''
\newblock {\em SSRN Electronic Journal}, 01 2022.

\bibitem{li2021learning}
Li~S. and et~al.,
\newblock ``Learning residue-aware correlation filters and refining scale for
  real-time uav tracking,''
\newblock {\em PR}, vol. 127, pp. 108614, 2022.

\bibitem{huang2019learning}
Huang Z. and et~al.,
\newblock ``Learning aberrance repressed correlation filters for real-time uav
  tracking,''
\newblock in {\em ICCV,2019}, pp. 2891--2900.

\bibitem{Zhang2022TrackingSA}
Zhewen Zhang and et~al.,
\newblock ``Tracking small and fast moving objects: A benchmark,''
\newblock in {\em Asian Conference on Computer Vision}, 2022.

\bibitem{Zhang2023TsfmoAB}
Zhewen Zhang and et~al.,
\newblock ``Tsfmo: A benchmark for tracking small and fast moving objects,''
\newblock {\em SSRN Electronic Journal}, 2023.

\bibitem{Liu2022GlobalFP}
Liu M. and et~al.,
\newblock ``Global filter pruning with self-attention for real-time uav
  tracking,''
\newblock in {\em BMVC}, 2022.

\bibitem{Zhong2022EfficiencyAP}
Zhong P. and et~al.,
\newblock ``Efficiency and precision trade-offs in uav tracking with filter
  pruning and dynamic channel weighting,''
\newblock in {\em FSDM}, 2022.

\bibitem{chen2020simple}
Chen T. and et~al.,
\newblock ``A simple framework for contrastive learning of visual
  representations,''
\newblock in {\em ICML}, 2020, pp. 1597--1607.

\bibitem{Park2020ContrastiveLF}
Park T. and et~al.,
\newblock ``Contrastive learning for unpaired image-to-image translation,''
\newblock in {\em ECCV}, 2020.

\bibitem{Zhang2021CrossModalCL}
H.~Z and et~al.,
\newblock ``Cross-modal contrastive learning for text-to-image generation,''
\newblock {\em CVPR}, pp. 833--842, 2021.

\bibitem{Li2022PairLevelSC}
Li~S. and et~al.,
\newblock ``Pair-level supervised contrastive learning for natural language
  inference,''
\newblock {\em ICASSP}, pp. 8237--8241, 2022.

\bibitem{Wu2021ProgressiveUL}
Wu~Wan J. and Chan A.B.,
\newblock ``Progressive unsupervised learning for visual object tracking,''
\newblock {\em CVPR}, pp. 2992--3001, 2021.

\bibitem{pi2022hierarchical}
Pi~Z. and et~al.,
\newblock ``Hierarchical feature embedding for visual tracking,''
\newblock in {\em ECCV}. Springer, 2022, pp. 428--445.

\bibitem{Pang2021QuasiDenseSL}
Pang J. and et~al.,
\newblock ``Quasi-dense similarity learning for multiple object tracking,''
\newblock {\em CVPR}, pp. 164--173, 2021.

\bibitem{Yu2022TowardsDR}
Yu~E. and et~al.,
\newblock ``Multi-view trajectory contrastive learning for online multi-object
  tracking,''
\newblock {\em CVPR}, pp. 8824--8833, 2022.

\bibitem{Li2021LearningRC}
Li~S. and et~al.,
\newblock ``Learning residue-aware correlation filters and refining scale
  estimates with the grabcut for real-time uav tracking,''
\newblock {\em 3DV}, pp. 1238--1248, 2021.

\bibitem{li2020asymmetric}
Li~S. and et~al.,
\newblock ``Asymmetric discriminative correlation filters for visual
  tracking,''
\newblock {\em FITEE}, vol. 21, no. 10, pp. 1467--1484, 2020.

\bibitem{Li2021EquivalenceOC}
Shuiwang Li and et~al.,
\newblock ``Equivalence of correlation filter and convolution filter in visual
  tracking,''
\newblock {\em ArXiv}, vol. abs/2105.00158, 2021.

\bibitem{Fu2021SiameseAP}
Fu~C. and et~al.,
\newblock ``Siamese anchor proposal network for high-speed aerial tracking,''
\newblock {\em ICRA}, pp. 510--516, 2021.

\bibitem{Cao2022TCTrackTC}
Cao Z. and et~al.,
\newblock ``Tctrack: Temporal contexts for aerial tracking,''
\newblock {\em CVPR}, pp. 14778--14788, 2022.

\bibitem{khosla2020supervised}
Prannay K. and et~al.,
\newblock ``Supervised contrastive learning,''
\newblock {\em NIPS}, vol. 33, pp. 18661--18673, 2020.

\bibitem{2015High}
Joao F. and et~al.,
\newblock ``High-speed tracking with kernelized correlation filters,''
\newblock {\em TPAMI,2015}, vol. 37, pp. 583--596.

\bibitem{danelljan2016adaptive}
Danelljan M. and et~al.,
\newblock ``Adaptive decontamination of the training set: A unified formulation
  for discriminative visual tracking,''
\newblock in {\em CVPR,2016}, pp. 1430--1438.

\bibitem{2017Learning}
Hamed G. and et~al.,
\newblock ``Learning background-aware correlation filters for visual
  tracking,''
\newblock in {\em ICCV}, 2017.

\bibitem{danelljan2017eco}
Danelljan M. and et~al.,
\newblock ``Eco: Efficient convolution operators for tracking,''
\newblock in {\em CVPR,2017}, pp. 6931--6939.

\bibitem{li2018learning}
Li~F. and et~al.,
\newblock ``Learning spatial-temporal regularized correlation filters for
  visual tracking,''
\newblock in {\em CVPR,2018}, pp. 4904--4913.

\bibitem{du2018the}
Du~D. and et~al.,
\newblock ``The unmanned aerial vehicle benchmark: Object detection and
  tracking,''
\newblock in {\em ECCV,2018}, pp. 375--391.

\bibitem{guo2021graph}
Guo D. and et~al.,
\newblock ``Graph attention tracking,''
\newblock in {\em CVPR,2021}, pp. 9543--9552.

\bibitem{zhang2021learn}
Zhang Z. and et~al.,
\newblock ``Learn to match: Automatic matching network design for visual
  tracking,''
\newblock in {\em ICCV,2021}, pp. 13339--13348.

\bibitem{Kim2022TowardsST}
Kim M. and et~al.,
\newblock ``Towards sequence-level training for visual tracking,''
\newblock {\em ArXiv}, vol. abs/2208.05810, 2022.

\bibitem{2022SparseTT}
Fu~Z. and et~al.,
\newblock ``Sparsett: Visual tracking with sparse transformers,''
\newblock {\em ArXiv}, vol. abs/2205.03776, 2022.

\bibitem{2016A}
Matthias M. and et~al.,
\newblock ``A benchmark and simulator for uav tracking,''
\newblock {\em FJMS,2016}, vol. 2, no. 2, pp. 445--461.

\bibitem{li2017visual}
Li~S. and et~al.,
\newblock ``Visual object tracking for unmanned aerial vehicles: A benchmark
  and new motion models.,''
\newblock in {\em AAAI,2017}, pp. 4140--4146.

\bibitem{wen2018visdrone}
Wen L. and et~al.,
\newblock ``Visdrone-sot2018: The vision meets drone single-object tracking
  challenge results,''
\newblock in {\em ECCV}, 2018, pp. 469--495.

\end{thebibliography}

\end{document}